\documentclass{KERauth}
\usepackage[algoruled, boxed, lined]{algorithm2e}
\usepackage{mathtools}
\usepackage{graphicx}
\usepackage{epstopdf}	
\usepackage{caption}

\begin{document}

\graphicspath{{pics/}}

\KER

\runningheads{Peters, C., Esfandiari, B., Zalat, M., and West, R.}{Behavior Cloning in OpenAI using Case Based Reasoning}



\title{Behavior Cloning in OpenAI using Case Based Reasoning}

\author{Chad Peters\affilnum{1},
Babak Esfandiari\affilnum{2},
Mohamad Zalat\affilnum{3} and
Robert West\affilnum{4}}

\address{\affilnum{1},\affilnum{4}Institute for Cognitive Science, Carleton University, Ottawa, Canada}

\address{\affilnum{2},\affilnum{3}Dept. Systems and Computer Engineering, Carleton University, Ottawa, Canada}


\begin{abstract}
Learning from Observation (LfO), also known as Behavioral Cloning,  is an approach for building software agents by recording the behavior of an expert (human or artificial) and using the recorded data to generate the required behavior. jLOAF is a platform that uses Case-Based Reasoning to achieve LfO.  In this paper we interface jLOAF  with the popular OpenAI Gym environment. Our experimental results show how our approach can be used to provide a baseline for comparison in this domain, as well as identify the strengths and weaknesses when dealing with environmental complexity.
\end{abstract}


\section{Introduction}

Behavioral cloning is the process by which an intelligent agent that has not yet learned the desired behavior can learn from another agent that is already capable of demonstrating the desired behavior in a target environment. The cloning process can take different forms, such as directly copying an encoded policy, or learning from direct observation. This paper describes our contributions to the study of behavioral cloning between agents using a form of Learning from Observation (LfO) through Case-Based Reasoning (CBR). We accomplished this through the use of the OpenAI Gym environment (Brockman et al., 2016) to test our agents.  

We approached this problem by using an agent already trained for near-optimal performance in various virtual environments, generating observable target behavior, and using a basic Case-Based Reasoning (CBR) approach to training a new agent with only the recordings of observable behavior as a reference. We use the Java Learning from Observation Framework (jLOAF), a framework that applies CBR to perform LfO (M W Floyd \& Esfandiari, 2011). 

As a result of this study, we show that jLOAF can be used to provide quickly with little effort an adequate baseline for the creation of agents in the OpenAI gym environment. Also, the interface that we built between jLOAF and gym will allow jLOAF to benefit to all the gym benchmarks and to be further improved. Our research design involved modeling the inputs and the outputs between the environment and the CBR system, so that learning agents can use the observed behavior of other expert agents already proficient in the target virtual environment, modify their own behavior based on similarity of state-action pairs, and resulting with a close approximation in terms of behavioral cloning. 

A cursory layout of the remainder of this paper is as follows; first we provide an overview of the related Learning from Observation (LfO) literature, and the cycle of working with the CBR system required for an agent’s information knowledge management. Next, we introduce jLOAF, a framework that supports learning in autonomous agents without using direct control by a human supervisor, including some previous examples of implementation in a few problem domains, and discussion of this framework is designed to be extended in its current form to adapt to future problem domains. The following section provides an introduction to the OpenAI Gym environment, and an architectural overview of our interface layer between the Java Virtual Machine (JVM), and Python Interpreter used by jLOAF and Gym, respectively. The experimental methodology includes how the two environments interface, generation of recorded observations used to build the base-cases, and measurements used to compare the performance and accuracy trade-off of the similarity and sampling methods. Last, we discuss the experimental results, where we evaluate the quality of our generated agents, and gradually improve our baseline results by some simple feature engineering and sampling. 


\section{Background}

There are a variety of ways that intelligent agents can learn new concepts or ideas, or improve on existing ones. Machine Learning paradigms can be described in terms of how the reward function is delivered to the learning agent. In Reinforcement Learning, the feedback comes directly from the environment, and is used to tune a policy function to predict the expected utility across multiple actions (Sutton \& Barto, 2012). In some cases, however, a reward function may not be directly available; it can be difficult to describe the proper function when either the human expert does not have personal experience beyond that of observation, or the problem domain reaches a level of complexity that is difficult to describe using a heuristic or programmatic approach. This can be seen in example of optimal behavior that can only be indirectly compared to the agent attempting to learn some function or trajectory given an observed state in the environment (Ontañón, Montaña, \& Gonzalez, 2014). In these types of scenarios, we can resort to a form of learning commonly known as Learning from Observation (LfO), where the actions to be modeled become the labels of the examples provided to the learner. 

\subsection{Learning from Observation}

Humans and non-human mammals exhibit the ability to learn from reinforcement from birth (Friedenberg \& Silverman, 2005), and as intelligent agents, still rely on some form of feedback (Russell \& Norvig, 2009), whether supervised (by others), or unsupervised (by ourselves) using a memory-based system of recall.

The concept of Learning from Observation (also known as Learning from Demonstration), provides an affordance to teaching intelligent agents by providing sample behaviors to learn from, and removing the requirement for direct intervention by the researcher. This can be done by presenting ideal examples of desired behavior to the learning agent, and through some means of encoding the storage and retrieval of these examples, the learning agent has the opportunity to compare its behavior against the optimal example for the purpose of self correction.

Argall et al.(Argall, Chernova, Veloso, \& Browning, 2009) argue that regardless if the environment is physical or virtual, one needs to encode the agent’s observation before it can be compared algorithmically, regardless of how the algorithm is represented. With some careful creation of a method to transform observations into a policy, and compare that policy against future actions (Ontanon, Mishra, Sugandh,\& Ram, 2008). 

An example implementation that satisfies these requirements has been implemented by Floyd and Esfandiari (Floyd \& Esfandiari, 2011) in the Java Learning from Observation  Framework (jLOAF). The current  version of jLOAF implements LfO through indeterminate inputs, feature selection and filtering, Case Base creation and pruning, and time-sensitive representation through a method called temporal backtracking (Floyd \& Esfandiari, 2011). One well-known paradigm of LfO is through the creation of state-action pairs, or cases, for use in case-based reasoning, as described in the following section.

\subsection{Case Based Reasoning}

CBR is an approach to problem solving that takes advantage of previously experienced situations in order to infer a probable solution to new experiences (Aamodt \& Plaza, 1994). This paradigm differs from other approaches to problem solving that rely solely on a general understanding of the problem domain, and instead leverages specific cases that can be reused in new ways and applied to new experiences.
 
Aamodt and Plaza describe a number of methods (Aamodt \& Plaza, 1994) that can be used to index, organize, retrieve, and utilize the information observed in the past, such as by \emph{explar} (finding the right class for an unclassified problem), \emph{instances }(combining cases to form concepts), \emph{memory }(reasoning through search), \emph{analogy }(using a different-but-similar domain), and typical \emph{case-base }(retrieving and adapting similar cases to new problems). Despite their subtle differences, all of these CBR methods can be represented as a cycle with four  main phases: \emph{Retrieve }cases that are the most similar to the current observation; \emph{Reuse }the case if it is a suitable match for the task at hand; \emph{Revise }(or create) the case to better reflect the currently observed; and \emph{Retain }the case for future use. In the next section, we describe how our methodology creates a case base through observing expert agents in different domains, and applies this reasoning cycle to each environment in question.


\section{Research Methodology}

The goal of this project was to evaluate a basic CBR approach to cloning the behavior of domain experts to quickly establish comparative baselines across multiple environments, and which may later be improved upon with more sophisticated reinforcement learning techniques.  In order to test this approach, we selected frameworks purpose-built to support Learning from Observation and Reinforcement Learning, the ability to establish measurements of state-action pairs through a sequence of time, and the ability to encode and transmit this information between the learning agent and each virtual environment. 

We selected the OpenAI  research platform as it provides a standardized interface for agents to observe, interact with, and receive feedback from a variety of virtual environments. OpenAI, described in more detail in the next section, has two modes: \emph{Gym }presents an interface for training and testing agents in a local environment, and \emph{Universe }for accessing cloud-hosted environments hosted remotely through a local proxy. Although the Gym testbed offers a reinforcement value for training reinforcement learning agents, our approach was to train and test an agent using only the observable behavior of an example (near-)optimal agent already trained using standard Reinforcement Learning techniques. 

We also selected the jLOAF framework, described in more detail in the next section,  to support and demonstrate the ability to interpret the behavior of expert agents, generate a case base that represents this behavior in the target environment, interact with the same environment using this case base, and measure the overall performance of various Reasoner classes provided by the framework in real-time. 

Our decision to couple jLOAF and Gym presented several architectural challenges due to the inherent differences between jLOAF and Gym, most notably the use of incompatible runtime environments. jLOAF uses the Java Virtual Machine, whereas Gym only supports Python . Despite these challenges, we created additional requirements to ensure the middleware system was developed according to sound software development practices. First, it must provide an extensible interface be-tween two disparate systems without sacrificing usability of either; Second, it must support both offline and online learning and communication between agent and environment; and finally, it must be forwards-compatible with future research projects based on OpenAI, such as the online Universe API used to interface with environments hosted remotely.

In order to properly evaluate agents operating in different environments with unique feature and action spaces, we designed a space-agnostic extension to abstract performance evaluation interface already available in the jLOAF framework so that each Case Base could be generated (based on logs from trained agents), stored, retrieved, and evaluated independent of the dimensionality of the observable environment space reported by Gym.

Our research project demonstrates the application of jLOAF to an open-ended Reinforcement Learning platform created by OpenAI, and specifically the Gym environment that is used for tuning and testing reinforcement learning agents. 


\section{An Overview of jLOAF}

The Java Learning from ObservAtion Framework (jLOAF) used in this study implements the general concepts of a CBR system through a collection of abstract classes that can be implemented by an agent to perform Learning by Observation in various environments. The overall architecture can be modified and extended; however all implementations have the following components in common: 

\begin{itemize}

\item
\textbf{Agent}: represents the central organizing construct that contains implementations of the others in order to observe the environment, reason about observations, select the most appropriate action, and perform those actions. The implemented agent class is what the main thread will instantiate and invoke. 

\item
\textbf{Input}: provides a way to encapsulate individual features that represent the environment. It supports both discrete and continuous variables, and can represent Atomic (single feature) and Complex (Atomic or Complex) representations in a recursive hierarchy.\\ 

\item
\textbf{Action}: can also use Atomic and Complex representations, and represents the outcome of the agent reasoning process, and to interact with the environment.\\
\item
\textbf{Similarity}: is the metric by which two Inputs are compared to each other. jLOAF provides similarity strategies for both Atomic and Complex inputs and actions, so these inputs can be compared for case retrieval.\\
\item
\textbf{Reasoning}: is the method by which the agent learns the behavior of the observed expert and predicting the next action for a given input. There are a number of built-in reasoners that use Machine Learning techniques such as Bayesian, Neural Net-work, Temporal Backtracking, and k-Nearest Neighbour. \\
\item
\textbf{Performance}: provides the template for evaluating how well an agent learns the target behavior and performs in new situations through 10-fold Cross Validation. Statistical libraries provide common measurements for comparison, such as Precision, Recall, and F-Score. \\
\item
\textbf{Filters}: can be used to tune agent performance through both feature selection, and case-base optimization. Feature selection can apply weights to the feature space based on perceived utility, or ignore them outright. Clustering of the case-base allows the reduction of the overall size by combining like-cases (which is especially important if the agent is to be able to perform in real time), whereas Sampling provides over- and under-sampling majority and minority classes, respectively, to deal with class imbalance as well as the reduction of the case base. \\

\end{itemize}


\section{The OpenAI Gym API}

The OpenAI project aims to provide a standard interface for agents to learn from and act upon a variety of virtual environments and problem domains. The main goal of the OpenAI project is to provide a common platform for researchers to compare and discuss novel reinforcement learning algorithms, and find a generalized solution to allow learning in a variety of domains. Our use of Gym for LfO is somewhat unique in that we are not trying to come up with an optimal agent that can beat humans at the task; instead, we wish to approximate and clone another agent, regardless of how good or bad the ex-ample may be by creating cases using the state-action pairs observed during execution of any other Gym agent. 

The Gym environment is supported in Python  2 and 3, works in both Linux and Windows (the authors have successfully used Gym on both platforms), and provides a standardized interface for each environment. A simple environment is instantiated in a python script  (as shown in Figure 1), and runs on the local machine of the Gym host. 


\begin{algorithm}
\KwIn{desired environment}
\KwOut{total reward}
\caption{Example Gym script}
import gym\;
initialize env\;
\While{env not done}{
	render environment\;
	observation = step (action)\;

}
\end{algorithm}

The Gym framework uses a standard agent-environment loop that steps through a new frame whenever the environment's step function is called, and returns a vector of four values:

\begin{itemize}

\item \textbf{Observation }(object): the state of the environment, represented as an array of double values. These values can represent any number of features, from the position or angle of an object, to a pixel on the screen. This representation is left up to the environment creator.\\

\item \textbf{Reward }(float): the reinforcement value used for an to learn in order to maximize the utility of each action. This value can take different ranges for the completion of each environment, as well as signal major events, such as entering a failed state, or achieving a checkpoint required for later success.\\

\item \textbf{Done }(Boolean): returns true if the environment has finished one round, otherwise always false.\\

\item \textbf{Info }(dictionary): a key-value collection that provides additional information about the state of the game. The OpenAI Gym standards do not allow agents to use this information in order to gain an advantage; rather it can be used by the re-searcher for development and debugging.\\

\end{itemize}

\subsection{Problem Classes in Gym}

The Gym platform divides the environments into problems subtypes, depending on a number of factors such as the complexity of the representation, the possible feature-action space, and the increasing degree of overall difficulty to put the agent into a “solved” state. Example problem classes include Search\& Optimization for text-based board games, Classic Control using a joystick or control pad, Atari games such as Breakout, and Box2D environments that can scale up for modern displays. 

\subsection{Action and Observation Spaces in Gym}

The Gym API also defines the concept of a space , that allows the calling agent to briefly interrogate the allowable actions for that environment, as well as the expected range for each feature in an observation. For example, the Lunar Lander environment will report a total of four allowable actions for each thruster, and the expected number range for features that describe the position, angle, and velocity of various dimensions. Feature spaces can be a standard unit vector represented as \emph{[-1, +1]}, or an infinite boundary represented as \emph{[-inf, +inf]}.

\subsection{Client-Server Model}

The jLOAF-OpenAI interface uses a client-server paradigm to allow communication between the jLOAF agent running in a Java Virtual Machine, and the Gym Environment running in a Python Interpreter VM. We are using Py4J  to provide the underlying communication framework. 

The Py4J package allows the jLOAF Agent and Gym Environment to communicate with each other through Inter-Process Communication (IPC) between the two virtual machines over TCP/IP, and an API for encapsulating the objects of the corresponding language constructs. This is accomplished by opening a socket and binding a port on each of the Client and Server Gateways for each call between Client and Server. The Py4J threading model  allocates a single thread for each call, and allows for event listeners and call backs if needed.

\begin{figure}
\centering
\captionsetup{justification=centering}
\vspace{1em}
\includegraphics[width=10cm]{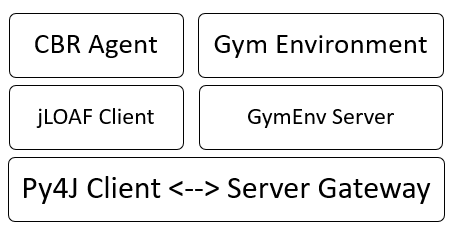}
\label{fig:Py4j_diag}
\caption{Client-Server interface between JLOAF agent and OpenAI Gym environments}
\end{figure}
 
The jLOAF client implements a Py4j.ClientServer bound to the GymEnv Python server entry point, and the GymEnv class defines the interface functions and parameters by which the Java client can remotely create, interrogate, reset, and shut-down any Gym environment pre-installed on the listening server.


\section{Experimental Design}

As defined in the Research Methodology, we wanted to capture near-optimal performance in a variety of environments to be used as the target behavior for observation and emulation. 

First, we used a variety of reinforcement learning-based agents as the teacher to be observed by our LfO agent. This allowed us to obtain an easily reproducible case base that is of a high quality. Some of the environments  come with a heuristic search-based reinforcement learning approach provided by the environment author. In these cases we relied on the heuristic agent as it performed more than adequately for a teacher. For the environments that did not have a solution, we used a stochastic gradient descent approach to reinforcement learning to generate a near-optimal policy for the teaching agent. These environments will be described further in this paper.

Next, the observed behavior of the above teachers is then used for Case-Base creation, and training of a new agent. In jLOAF, Case-Base creation is done by mapping the input-action space through the creation of atomic state-action pairs. These cases are compared against future observations made in similar environments to evaluate the accuracy of the cases retrieved from the case base. A simple, yet effective method of comparing the similarity of features is k-Nearest Neighbour (kNN). One could argue that given the simplicity of the observation space in Gym that is used to calculate a similarity functions, we could have directly implemented a simple k-Nearest Neighbor (kNN) function instead of using a framework running in a separate environment. However, we decided to turn a potential obstacle into a challenge to overcome for two reasons: First, the jLOAF library provides an extensive framework that supports any combination of atomic and complex state-action pairs, and scales well to large and complicated environments. This is important for the support of future environments that may compound on existing observation spaces. Second, the jLOAF library provides a test suite to compare different learning strategies across a range of statistical measurements. This is important if we wish to run combinations of offline and batch experiments to compare the time and accuracy trade-offs between different pre-processing and learning strategies. In this study we use a Euclidean distance to calculate the similarity between between Atomic and Complex inputs.

Last, the performance testing methodology comprises a number of scenarios to evaluate the overall accuracy between three elements: \emph{Environments} in Gym to provide varying levels of complexity; \emph{Reasoners }to train and predict the agent; and \emph{Filtering }to optimize the size of the case base.

The Gym toolset includes a variety of environment classes that can be used to train and test reinforcement agents of varying complexity and sophistication. Environment classes include algorithmic text, classic control problems found in machine learning textbooks, 2D and 3D physics and robotics simulators, and even some first-generation Atari  games. The chosen environments and their classes described in the next section were chosen and tested to generate supporting enrichment data to describe how the agents learn and perform.

\subsection{Classic Control Environments}

The classic control environments were selected as they provide Complex Inputs and Atomic Actions in a real-time environment as well as providing a reasonable starting point for testing.

\subsubsection{Cart Pole}

The Cart Pole game is presented as a vertical pole balancing on a cart that can move \emph{left} and \emph{right} on a horizontal axis. The purpose of this challenge is to keep the pole in an upright position for at least 200 game steps without letting it fall to either side more than 15 degrees from center.\\ 

\begin{figure}[h]
\centering
\captionsetup{justification=centering}
\vspace{1em}
\includegraphics[width=7cm]{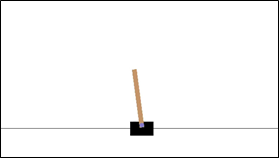}
\label{fig:cartpole_env}
\caption{Cart Pole Gym Environment}
\end{figure}

Each game step the environment is presented as an array of four (4) elements, summarized in the table below:

\begin{table}[h]
\centering
\label{tab:cartpole_tab}
\caption{Cart Pole Observation Space}
\begin{tabular}{ |c|c|c|c| } 
\hline
Num & Observation	& Min & Max \\
\hline
0 & 	Cart Position & -2.4 &	2.4 \\
\hline
1 & 	Cart Velocity & -Inf & 	Inf \\
\hline
2 &	Pole Angle &	-41.8 &	41.8 \\
\hline
3 & 	Pole Velocity At Tip &	-Inf & 	Inf \\
\hline
\end{tabular}
\end{table}

The environment accepts one of two possible actions, representing the left and right controls to move the cart in the desired direction. 

\subsubsection{Mountain Car}

The Mountain Car environment is presented as a car sitting at the bottom of a valley, with the task of building enough momentum to reach the goal at the top of the hill. This is considered a slightly more challenging task, as learning agents must learn to first move away from the goal before they have enough momentum to reach the top of the other side.\\

\begin{figure}[h]
\centering
\captionsetup{justification=centering}
\vspace{1em}
\includegraphics[width=7cm]{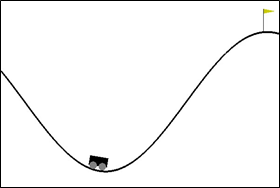}
\label{fig:mc_env}
\caption{Mountain Car Gym Environment}
\end{figure}

Each game step, the Mountain Car environment is presented as an array of two (2) elements, representing the distance from the goal, and the current velocity. 

\begin{center}
\begin{table}[h]
\centering
\label{tab:mc_tab}
\caption{Mountain Car Observation Space}
\begin{tabular}{ |c|c|c|c| }
\hline
Num & Observation	& Min & Max \\
\hline
0	& position	& -1.2 & 	0.6\\
\hline
1&	velocity	& -0.07	& 0.07\\
\hline
\end{tabular}
\end{table}
\end{center}

The environment accepts one of three possible actions; namely \emph{move left},\emph{move right}, or \emph{slow down}.

\subsection{2D Box Environment}

A more advanced example is the “Box2D” environment, as it provides a considerably more complex observation space than Classic Control, and allows for higher difficulty by forcing agents to start in a randomly generated starting point. 

\subsubsection{Lunar Lander}

The Lunar Lander environment simulates a rudimentary rocket ship in two dimensions tasked with safely landing in a specified area without running out of fuel. This environment randomly generates the terrain and starting position, and allows some level of customization through altering a random seed value.\\ 

\begin{figure}[h]
\centering
\captionsetup{justification=centering}
\vspace{1em}
\includegraphics[width=7cm]{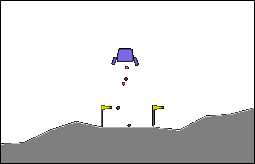}
\label{fig:lunar_env}
\caption{Lunar Lander Gym Environment}
\end{figure}

The observation space consists of eight (8) elements that represent the lander location, trajectory, spin, and whether or not one of the two legs is touching the landing pad. The player is penalized for “bouncing” off the pad, encouraging behaviors that land in one shot. The feature range makes normalization especially difficult, as the reported range is infinite for all values. We addressed this issue by generating a distribution for each feature based on sampling multiple rounds of the heuristic lander. The elements of the environment as follows:
 
\begin{center}
\begin{table}[h]
\centering
\label{tab:lunar_tab}
\caption{Lunar Lander Observation Features}
\begin{tabular}{ |c|c|c|c| }
\hline
Num	&Observation	&Min	&Max\\
\hline
0&	x position	&-Inf	&Inf\\
\hline
1	&y position	&-Inf	&Inf\\
\hline
2	&x velocity	&-Inf	&Inf\\
\hline
3	&y velocity	&-Inf	&Inf\\
\hline
4	&angle	&-Inf	&Inf\\
\hline
5	&angular velocity	&-Inf	&Inf\\
\hline
6	&Left leg contact	&0	&1\\
\hline
7	&Right leg contact	&0	&1\\
\hline
\end{tabular}
\end{table}
\end{center}

There are four (4) actions available to the agent, which are: \emph{do nothing, fire left engine, fire main engine }and \emph{fire right engine}. Firing the left and right engine rotates the lander to the right and left respectively, while the main thruster pushes the lander in the direction of orientation. 

\subsection{Agent Evaluation}

We selected the K-Nearest Neighbour family of similarity strategies for the purpose of case retrieval, since they map well to the state-based input-action pairs presented by the environment. This is important as one of the main contributions of this research is providing a baseline that is both simple to measure, easy to replication, so future researchers using our approach can compare and contrast more sophisticated and effective LfO strategies.

In order to compare the learning strategies against each other, we used Cross Validation with 10 slices of pre-generated results from expert agents in each domain, each slice containing 2000 cycles of state-action sets. First, we compute the F-score across all classes of actions within the environment using the standard formula:

$$F_i=\frac{2 x precision x recall_i}{precision_i+recall_i }$$
such that
$$precision_i=\frac{c_i}{t_i}$$
and
$$recall_i=\frac{c_i}{n_i}$$

where $c_i$ is the count of matches between the known and generated actions, $t_i$ is total number of times an action is generated, and $n_i$  is the number of times the action should have been generated. The Global F-score can then be computed as:

$$F_{global}=\frac{1}{A}\sum_{i=1}^{A}F_i$$

We use the Global F-score to summarize the accuracy of all actions across all observations in the environment. See (Floyd\& Esfandiari, 2008) for various applications of Global F-score to atomic and complex inputs. 

To obtain the best possible results, we first varied the parameter $k$ to find its optimal value. We then normalized the data to ensure that one feature doesn’t overwhelm the others in the calculation of the similarity. Finally, we evaluated the use of sampling. The sampling algorithm we used is the condensed nearest-neighbor rule (Hart, P., 1968). In this technique, samples are provided to the learner one by one, and only samples that are guessed incorrectly are kept. In this way, only the samples that approximate boundaries between labels are kept, and so we not only reduce the number of samples for real time performance reasons, but we also deal with class imbalance, by oversampling under-represented data points, and under-sampling overly represented data points.

Thus, each combination of environment, reasoner, and filter was run to calculate the mean and standard deviation of Accuracy, Recall, Precision, and F-Score. The F-Scores were then averaged across case sizes to provide a Global F-Score used to asses each agent configuration over the space of all possible actions given the feature representations provided.


\section{Experimental Results}

First we report on the effect of varying $k$ in each environment by gradually increasing $k$ from 1 through 5, 10, 15, 20, 30, and 50 to see the overall effect on agent accuracy.

\begin{figure}[h]
\centering
\captionsetup{justification=centering}
\vspace{1em}
\includegraphics[width=10cm]{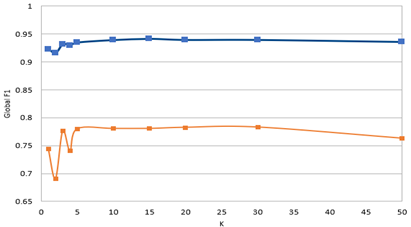}
\caption{Accuracy by varying k in Cart Pole}
\label{fig:vary_k_cartpole}
\end{figure}

We note that increasing the value of k in the Cart Pole environment (Figure \ref{fig:vary_k_cartpole}) increases the accuracy of the LfO agent from \emph{0.91 }to \emph{0.95}, and immediately plateaus around $k=10$ both with and without filtering.

\begin{figure}[h]
\centering
\captionsetup{justification=centering}
\vspace{1em}
\includegraphics[width=10cm]{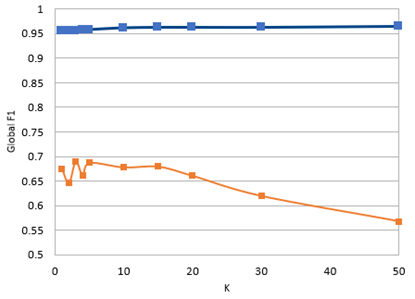}
\caption{Accuracy by varying k in Mountain Car}
\label{fig:vary_k_mc}
\end{figure}

\begin{figure}[h]
\centering
\captionsetup{justification=centering}
\vspace{1em}
\includegraphics[width=10cm]{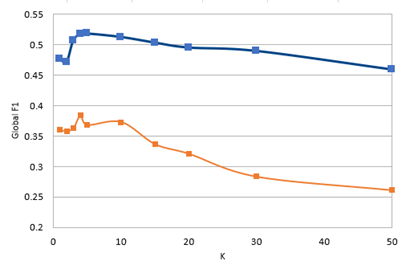}
\caption{Accuracy by varying k in Lunar Lander}
\label{fig:vary_k_lunar}
\end{figure}

The Mountain Car environment (Figure \ref{fig:vary_k_mc}) and Lunar Lander environment (Figure \ref{fig:vary_k_lunar}), however, revealed that increasing the value of k beyond 10 has much greater (and detrimental) effect in combination with filtering the case base in both environments, which are somewhat more complex and subject to the class imbalance problem. 

We thus came to the conclusion that a value of $k$ between 5 and 10 is optimal for \emph{most} games in the simpler problem classes.

\subsection{Sampling Effects}

All but the most trivial games such as Cart Pole may require the agent to use one set of actions far more often than others to reach the objective; this is a common problem that introduces classification imbalance during generation of the case base, resulting in agents that are biased toward specific actions. We deal with the issue of class imbalance by reduction by sampling the cases using ....

The default Cart Pole environment essentially splits the cases on either side of a perfectly upright (vertical) pole, and therefore makes intuitive sense that the observed state-action pairs would be proportionately represented. In a sample of 20,000 cases, we observed 10133 and 9867 (51\% to 49\%) thus confirming our intuition. 

\begin{figure}[h]
\centering
\vspace{1em}
\includegraphics[width=10cm]{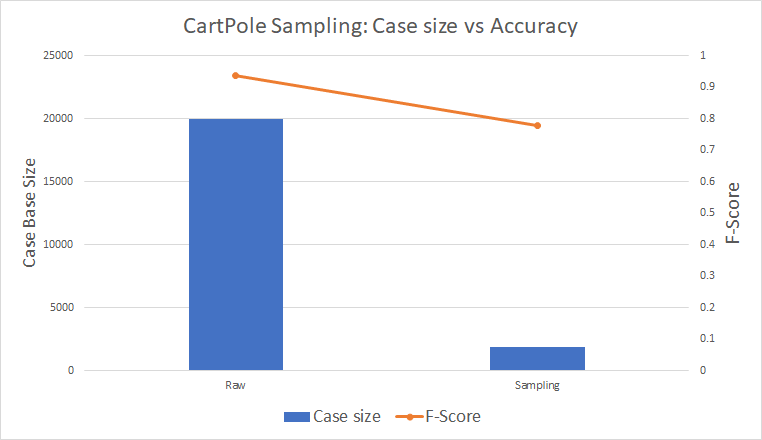}
\caption{Cart Pole Sampling Effect on Case Base Size and Accuracy}
\label{fig:sampling-cp}
\end{figure}

The mean case base size after sampling in Cart Pole environment is 1880. This is much lower compared to the 18000 cases in the original case base, an 89.6\% decrease in case base size (Figure \ref{fig:sampling-cp}).

However, the maximum Global F1 measure was also reduced from 0.941 to 0.783. This is a reasonable tradeoff considering the 89.6\% decrease in case base size which improves the performance. The optimal $k$ was 15 before sampling and then around 20 after sampling, this difference is not significant enough as the global F1 score of $k=15$ is very close to that of $k=20$ after sampling.

In the Mountain Car environment, the case base size was reduced by 97.6\% (from 18000 to a mean of 438) which is a substantial reduction (Figure \ref{fig:sampling-mc}). On the other hand, the global F1 score dropped to 0.691 from 0.965 after sampling (28.4\% drop in global F1 score, Figure \ref{fig:sampling-mc}). 

\begin{figure}[h]
\centering
\vspace{1em}
\includegraphics[width=14cm]{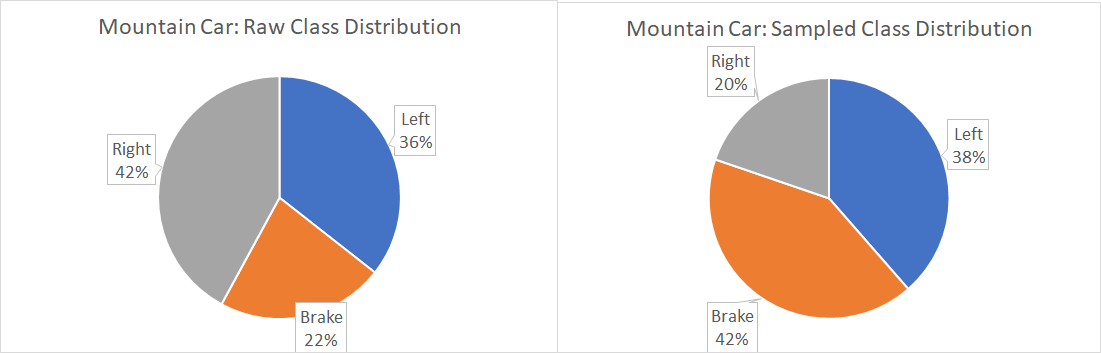}
\caption{Mountain Car Action Space Class Distribution}
\label{fig:mc-classdist-raw-sampled}
\end{figure}

\begin{figure}[h]
\centering
\vspace{1em}
\includegraphics[width=10cm]{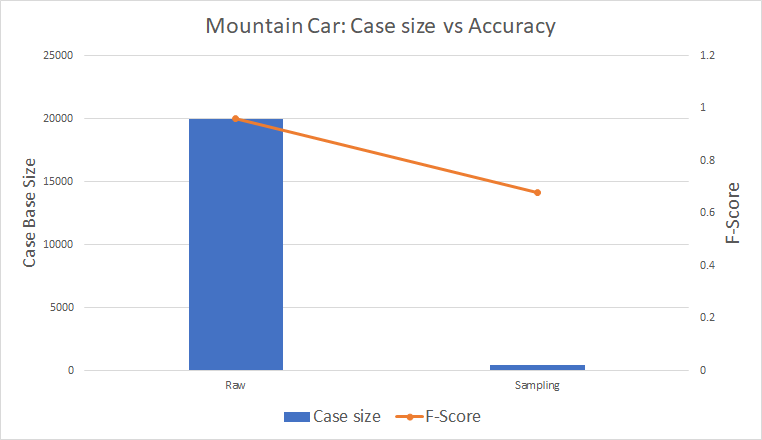}
\caption{Mountain Car Sampling Effect on Case Base Size and Accuracy}
\label{fig:sampling-mc}
\end{figure}

This drop in the F1 measure is much larger than the one observed in Cart Pole and may not be worth the trade-off for a smaller case base. The optimal $k$ was found to be at 3 after sampling, with the optimal $k$ being at 50 before sampling. Here there is a significant change in the optimal $k$ after applying sampling; this is important as it also helps with real-time performance, as calculating the majority action is faster for smaller values of $k$ . 

\begin{figure}[h]
\centering
\vspace{1em}
\includegraphics[width=14cm]{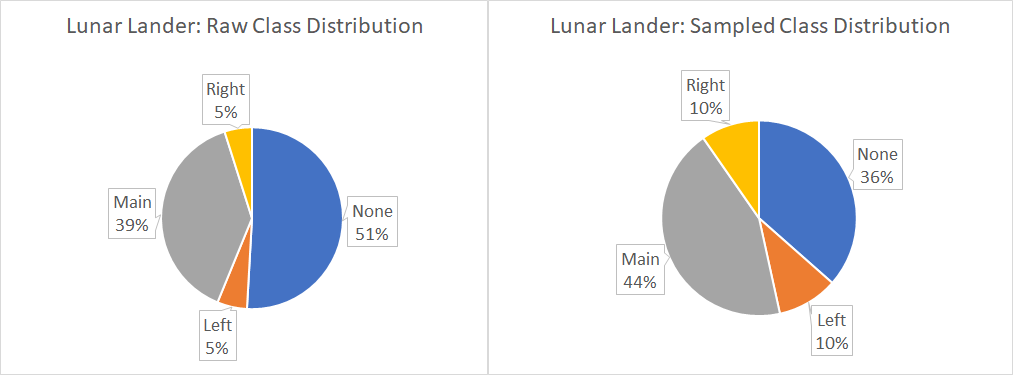}
\caption{Lunar Lander Action Space Class Distribution}
\label{fig:sampling-ll}
\end{figure}

As for Lunar Lander, the algorithm's performance was poor due to the environment's randomization (Figure \ref{fig:ll-classdist-raw_sampled}). The global F1 score was 0.518 before sampling and 0.384 after sampling (a 25.9\% decrease in the F1 score). However the case base size was reduced by 74.9\% (from 18000 to 4520 cases). This is still not a very acceptable trade-off given that the global F1 score was already very low.


The optimal $k$ did not significantly change before and after sampling ($k$ of 5 before sampling and $k$ of 4 after sampling).

\subsection{Visual Observations}

In addition to the qualitative methods as discussed, we want to judge the effectiveness of an agent based on approximation to the example (teacher) agent, as well as how human-like the learning behavior. Repeated observations of the agent operating in Gym environments produced some unusual results, all depending if the Gym environment uses a random seed to determine where the agent is situated when the environment is started. 

For example, the Cart Pole and Mountain Car environments will start off in the same position, with only minor differences in whether the pole is leaning slightly to the left or right. In cases of sampling and normalizing the range of observed teaching behavior, the agent was able to (eventually) self-correct. However, due to the near-optimal performance of the teaching agent, the learner did not have a chance to observe situations that forced the example agent to recover from a near-disaster; in these situations, forcing the teaching agent into random positions to recover from can be used to some degree of effectiveness to learn how to recover. Running the learner in simpler environments using a large case base size produced Global F-Scores as high as 0.98. In this scenario, the learner might be able to fool a human observer into thinking it is the example agent. 

When moving on to more complex environments with a random seed, the learner behavior diverges from the examples with a lower chance of recovery. For example, the Lunar Lander environment not only randomizes the terrain and landing target, but also supports a difficulty rating; even the easiest default setting assigns the lander starting locations, Cartesian velocities, and angular momentum (spin) that can be difficult for even human players to recover from. In situations when the agent started off in an ideal position and orientation as the example agent, the learner was able to land the craft in a similar fashion; however, when the learner had to recover from difficult scenarios, the outcome was almost always failure in completing the scenario, despite the successful examples demonstrated by the example agent. We tried to counter this problem by increasing the difficulty to generate a wider variety of starting positions for the teacher to adapt to, however, this did not increase the performance of the learning agent, even when the difficulty was turned down. This problem might be solved by oversampling the state space that a learner does not have the opportunity to observer in more complex environments, or providing additional simulation time for the learner to generate cases of successful recovery from a variety of scenarios. 
Additional work in this area is required to properly explain this phenomenon, and may include techniques such as Active Case Base generation (Michael W Floyd\& Esfandiari, 2009)  as a solution to these issues. 


\section{Conclusion}

This study provided a contribution to the study of Case-Based Reasoning and Learning from Observation by evaluating a basic CBR approach to behavioral cloning using the popular OpenAI Gym framework. We accomplished this by demonstrating a practical application of the jLOAF framework to an existing community-driven research platform, and the creation of a working interface that allows jLOAF users to leverage OpenAI technologies, while overcoming various technical hurdles to enable distributed communications. This research provided a comparison and analysis of the effects of optimzing the k-Nearest Neighbour similarity metric, Sampling on the overall accuracy across various environments, and generating a series of baseline examples for future comparison.

Future work in this domain may involve an extension of the testing framework to encapsulate some of the environment-specific requirements such as 3rd party libraries for each environment, and possibly porting the jLOAF system to Python to natively work with OpenAI Universe proxies

\end{document}